\documentclass[conference]{IEEEtran}
\usepackage{amsmath}

\usepackage{cite}
\usepackage{xcolor}
\usepackage{tikz}
\usetikzlibrary{arrows,shapes,snakes,automata,backgrounds,calc,petri,arrows,shapes,chains,matrix,positioning,scopes,automata,shapes.misc}
\usepackage{dblfloatfix}

\begin{document}
\title{Spiking Neural Algorithms for \\ Markov Process Random Walk}

\author{
\IEEEauthorblockN{William Severa, Rich Lehoucq, Ojas Parekh, and James B. Aimone}
\IEEEauthorblockA{Center for Computing Research\\
Sandia National Laboratories\\
Albuquerque, NM  87122\\
Email: wmsever@sandia.gov, rblehou@sandia.gov, odparek@sandia.gov,  jbaimon@sandia.gov}
}

\maketitle

\begin{abstract}
The random walk is a fundamental stochastic process that underlies many numerical tasks in scientific computing applications.  We consider here two neural algorithms that can be used to efficiently implement random walks on spiking neuromorphic hardware.  The first method tracks the positions of individual walkers independently by using a modular code inspired by the grid cell spatial representation in the brain.  The second method tracks the densities of random walkers at each spatial location directly.  We analyze the scaling complexity of each of these methods and illustrate their ability to model random walkers under different probabilistic conditions.
\end{abstract}

\IEEEpeerreviewmaketitle

\section{Introduction}

The classic random walk, a stochastic process, underlies many numerical computational tasks.  The random walk is a direct reflection of the underlying physical process and models Brownian Motion, among other processes.  Random walks have found myriad applications across a range of scientific disciplines including computer science, mathematics, physics, operations research, and economics~\cite{MASUDA20171}.  For instance, the treatment of ionic movements as a random walk process is critical to deriving Nernst-Plank dynamics for ions in understanding the biophysics of neurons~\cite{johnston1994foundations}.  Additionally, random walks are also used in non-physics domains, such as financial option pricing~\cite{black1973pricing} and ecology~\cite{codling2008random}. 

Random walks are typically straightforward to implement, and can be computationally appealing in high dimensional domains that are ill-suited for other numerical approaches.  Because they are typically used to independently sample a population, simulations of many random processes are easily distributed across a parallel machine; with each computational core responsible for a distinct process.  However, the utility of multi-core systems for multi-agent models such as random walks is still limited in many applications~\cite{chuang2013parallel}.  Most simulations that utilize random walks to statistically arrive at a solution require the aggregation of a population of walkers before any conclusions can be made.  Aggregation of walkers' behavior requires both time and energy, and the difficulties are exacerbated when scaling to large systems (e.g.~terascale, petascale, exascale).  Thus while the walkers themselves are easily parallelized, the overall simulation is still constrained by the integration of information across the population.

Similar to other heavily parallel domains, parallelization via GPUs can out-perform CPUs for random walks, seeing roughly an order-of-magnitude performance gain~\cite{vanAntwerpen2011improving}.  However, these gains are offset by increased power consumption which renders the benefits dubious in an HPC setting.  Additionally, the aggregation challenge is unimproved from the CPU implementation.

Neuromorphic hardware presents a compelling architecture to consider for an energy efficient implementation of random processes.  In the ideal, a neuromorphic platform can be viewed as an incredibly large parallel architecture, albeit one with very simple processors (i.e., the neuron)~\cite{severa2016spiking}.  These novel and upcoming platforms potentially offer dramatic improvements in performance-per-Watt~\cite{merolla2014million, diamond2016comparing, hasler2013finding}. Additionally, we hypothesize that neuromorphic platforms that leverage spiking neurons, such as the leaky integrate-and-fire (LIF) neuron, and have inherent capability for probabilistic sampling, such as either stochastic synapses or probabilistic thresholds, may offer compelling advantages for modeling a random walk process.  These stochastic components have been used for computation in other applications~\cite{buesing2011neural} and can be instantiated in hardware such as on IBM's TrueNorth chip~\cite{merolla2014million}.

This paper describes two spiking neural circuits for simulating random walkers.   We then analyze these models in the context of emerging neuromorphic computing architectures, such as the Intel Loihi chip~\cite{davies2018loihi} and the ARM-core based Manchester SpiNNaker platform~\cite{furber2013overview}.  We note that the approach taken here for modeling stochastic processes relies on relatively small circuits with very precise use of stochastic events, whereas an alternative approach to modeling stochastic inference consists of more dynamical population models of neurons~\cite{buesing2011neural}.

\section{Random Walk Model}

Consider a system, \textit{S}, that consists of a mesh of discrete locations.  For simplicity we will consider the case where the mesh is an lattice of $N$ grid points along each of $D$ dimensions, although in practice a lattice is not a requirement.  Within \textit{S} is a population that evolves through a random walk process that is suitable to model as a population of independent particles, such as a diffusion process where each particle moves through space according to a Brownian motion evolution.  We consider only the case where each particle is independent without interactions.

If a simulation models $K$ independent particles, then the average position of the $K$ particles approaches the expected value of the population at a rate of $O(1/ \sqrt{K})$ as a consequence of the central limit theorem.

\section{Neural model of random walk}

In this paper, we consider two neural circuit approaches to modeling a population of random walkers.  The first is a neural circuit to perform the conventional task of modeling each walker independently as it moves over a space, which we call the \textit{particle method}.  The second approach is a neural circuit that tracks the number of particles at each given location in a simulation.  This method effectively tracks the distribution of particles over the whole space, as we refer to it as the \textit{density method}.  The following two sections describe the motivation and circuits used to compute these respective methods, and the subsequent section describes the simulations performed to illustrate these approaches efficacy.

Importantly, as one of the motivating features of spiking neuromorphic hardware is its potential advantages in energy consumption, we consider here not only the required neuron resources and time to simulate these models, but also estimate the energy consumption of these models, for which number of spiking events is generally considered a first-order proxy.

\subsection{Particle models}

The most straightforward approach to modeling a random walk is to commit a subset of neurons to modeling each particle independently.  A simple neural algorithm for a particle consists of three parts: the \textit{stochastic process}, which determines what random action is taken, a \textit{spatial location}, which tracks the location, and an \textit{action circuit}, which updates the location based on the output of the stochastic process and any boundary conditions, if relevant.  

In most implementations, the dominating neuron cost for simulating individual walkers will be the spatial location.  Even if particles are relatively restricted in their local movements, each particle circuit must be able to represent all spatial locations that are relevant for the simulation.  Thus, if space (i.e., number of neurons) is the primary consideration, a compact code, such as a binary representation is well suited, as it requires only $O(D\cdot \log N)$ neurons to represent space.  However, a binary code is non-trivial to update using neurons, and the average activity of the network is relatively dense.   Alternatively, a unary code --- where one neuron represents each spatial location --- can be highly energy efficient (only one spike required to communicate location) and straightforward to update, albeit spatially impractical (requires $O(N^D)$ neurons to represent space).  

Here, we present a neural algorithm that lies between these extremes, offering a compromise between a binary and unary representation of space.

\subsubsection{A modular spatial code balances compactness and energy efficiency}

\begin{figure}[h]
\includegraphics[width=3.4in]{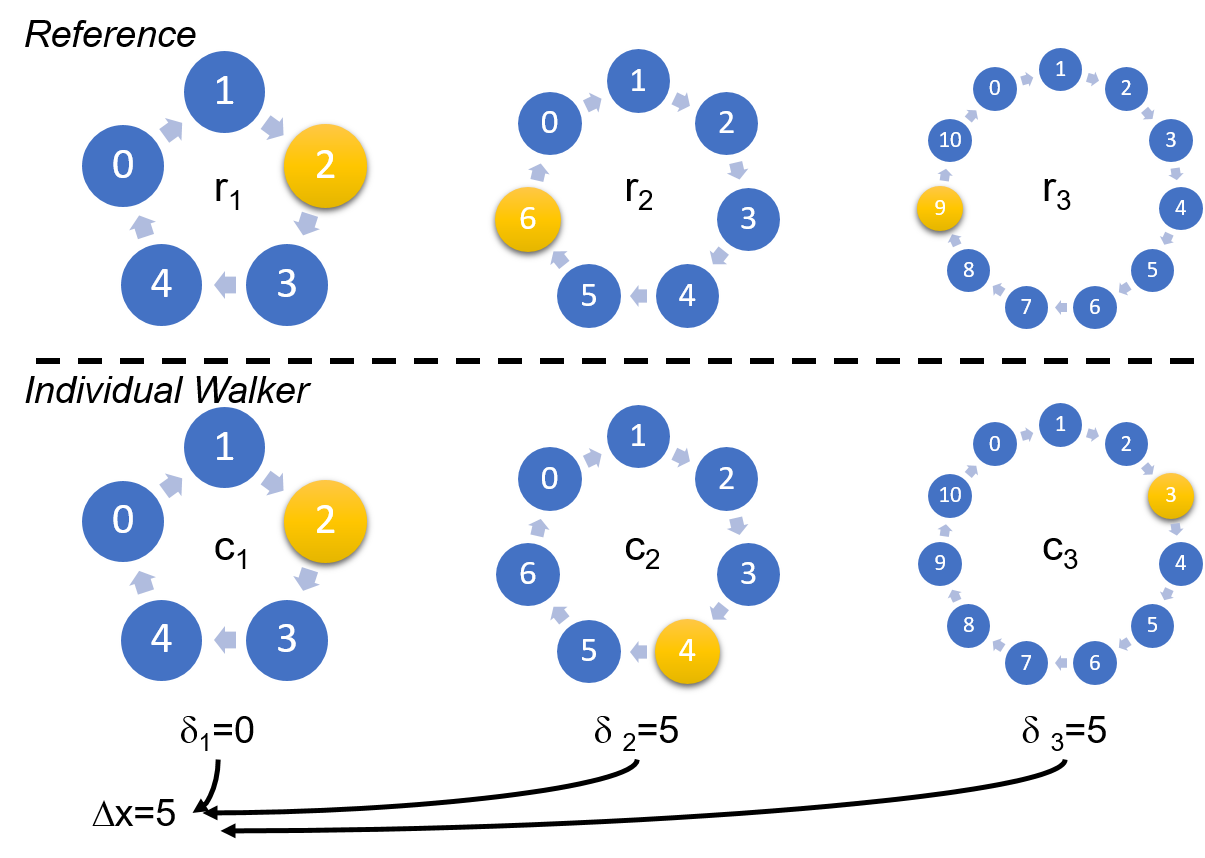}
\caption{Illustration of modular oscillator code.  Top panel shows reference oscillators, with rings of size 5, 7, and 11.  Individual walker have equivalent sized rings.  When not moving, all oscillators progress forward at same speed, however if a particle takes a random step, its individual rings will advance relative to the reference.  The position of each walker can then be decoded from these modular differences}
\label{ring_code}
\end{figure}

One potential model that lies between unary and binary is a modular code, also known as a residue numeral system. Our approach to implementing a modular code is shown in Fig.~\ref{ring_code}.  This model is inspired by a model for grid cells in the entorhinal cortex brain, which has been shown to have very high capacity for spatial locations relative to the more unary-like place cells in the hippocampus \cite{sreenivasan2011grid}. 

\begin{figure}[h]
\includegraphics[width=3.4in]{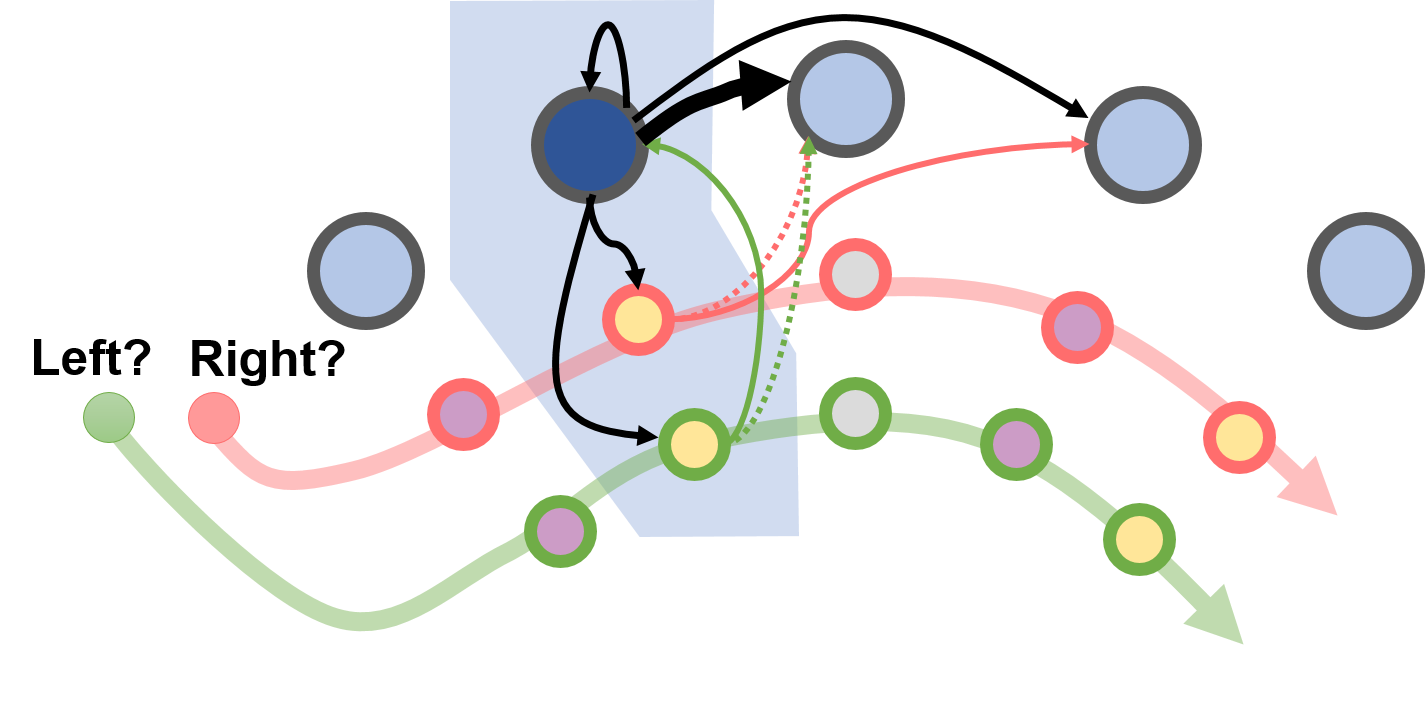}
\caption{Schematic of spiking neural circuit to construct ring osccilators with update capabilities.  Principal ring neurons are shown in blue, with the active neuron in dark blue.  Update neurons are below the ring neuron, and can be reused assigned neurons are at least three apart.  Bold solid line between adjacent ring neurons is weight $0.5$ and delay $2$; solid lines between other ring neurons are weight $0.5$ and delay $2$; solid lines to update neurons are weight $0.5$ delay $1$; dotted lines $-0.5$ and delay $1$; and movement source inputs are weight $0.5$ and delay $1$}
\label{ring_circuit}
\end{figure}

For each dimension, the particle circuit will have $M$ ring oscillators, each with a unique prime number of neurons, $C_i$ for $i \le M$, with states at time $t$, $c_i(t) \text{ for } i \le M$ with the combined state represented by the vector $C(t)=[c_1(t), c_2(t), ..., c_M{t}]$, where each state is the integer index of which neuron is active in each ring. This provides the circuit with $C_M=\prod{C_i}$  possible states. For example, consider a particle with $M=3$ and $C_1=3, C_2=5, \text{ and } C_3=7$, then the particle's spatial code would have $C_M=105$ possible states.  

To implement the random walk in neurons, we consider the case where a position $x$ is encoded by the offset between the particle's state vector $C$ and an equivalently sized reference population, $R$, which consists of rings of the same size.  At each time-step, for the state of each ring oscillator in the reference and particles advances by one, 

\begin{equation}
	c_i(t+1)=
	\begin{cases}
		c_i(t)+1, & \text{if } c_i(t)<C_i\\
		0,        & \text{otherwise}
	\end{cases}
	\label{modular_add}
\end{equation}

The position, $x$ is then generated from $C$ and $R$ by subtracting the two states.  For each oscillator, a difference

\begin{equation}
	\delta_i=(c_i-r_i)\mod C_i
	\label{offset}
\end{equation}

is computed, from which we know, by the Chinese Remainder Theorem, that the position, $x$, can be decoded. (One useful reference may be pages $873$--$876$ in \cite{cormen2001introduction}.)  One extension of residual codes such as these is that addition and multiplication involving $x$ can be performed by the equivalent modular arithmetic operation on each of the component rings.  Therefore, a change of $\Delta x$ in the position of a walker can be represented by adding $\Delta x$ to each of the states $c_i(t)$.

Structuring a neural circuit to advance a ring oscillator continuously is straightforward, with each ring of $C_i$ size being comprised of $C_i$ LIF neurons connected in a ring configuration, with the synaptic efficacy being sufficient to drive the downstream neuron to fire.  However, a non-obvious circuit is necessary to reliably speed up or slow down the oscillators if the random walk moves the location.  Fig.~\ref{ring_circuit} shows one circuit solution that uses spike delays to `add' and `subtract' to the position of the ring by one using spike delays.  The integrate-and-fire neurons in this circuit all have a spiking threshold of 1, a reset value of 0, and immediate decay (i.e., a time constant of 0).  In this implementation, a ring neuron at location \textit{i} is connected to the neuron at \textit{i+1} with a weight $1$, and to the neuron at \textit{i+2} and to itself with weight $0.5$.  Each of these ring connections has a delay of $2$.   With this setup, without any other inputs the ring will advance by one state every $2$ clock cycles.  

A secondary circuit is then placed on all rings of a walker to advance or stall the circuit (thus generating an offset relative to the reference).  We consider here the case where the particle has three potential movements (`left', `right', or `stay'); with a source neuron for each direction using a stochastic threshold or synapse to determine whether to move in one direction or not and communicating that action to each of the particle's rings for that dimension.  The currently active ring neuron, $i$, sends an input of weight $0.5$ and delay $1$ to both of its respective update neurons.  In the case where the circuit is advanced (labeled `right' in Fig.~\ref{ring_circuit}), all positive update neurons get a $0.5$ input as well, allowing the appropriate positive update neuron to fire.  That neuron then sends a +$0.5$ to the $i+2$ ring neuron and a -$0.5$ to the $i+1$ ring neuron, effectively shifting the ring forward by $1$.  The negative update is similar, except for driving the source $i$ neuron rather than the $i+2$ neuron.     

Importantly, because the rings are only locally activated and impact up to two ring neurons away, these update neurons can be reused every three ring neurons.  Ultimately, this means either four or five pairs of update neurons are required, because there are a prime number of ring neurons.

The dynamical representation of position as the offset of these oscillators confers several advantages.  First, it is consistent with the transient state of neurons.  Rather than a neuron having to self-activate to maintain a state, the ring simply evolves at a steady rate when position is not changing.  Second, it allows updates to be more efficiently implemented.  When there is a random movement of the particle, in whatever dimension is being considered, the particle's rings are in unison accelerated or decelerated by one.  The use of a common reference for all particles also allows changes in the frame-of-reference to be efficiently accounted for as well --- a simple shift in the reference state is the equivalent of shifting all the particles in unison.  This may be of use in models where an observer of a random walk is itself in motion.  Similarly, because the reference is used only in the decoding of position, it is possible to have multiple references, or to readily compute the distance between particles without using a reference at all.

\subsubsection{Complexity of oscillator particle model}

Each walker for the above model requires $2+\sum{C_i+2*(3+C_i \mod  3)}$ neurons and $\sum{9*C_i+2*(3+ C_i \mod 3)}$ synapses.  Only one spike is required per ring, for $M$ total, when there are no updates, and  $M+1$ additional spikes required for an update.

There is a global cost as well, with an additional set of rings for the reference position (although unless the reference position is also in motion, update neurons would not be required).  Each dimension would consist of its own rings.

This model presents a useful trade-off between a dense code, with lots of rings, and a sparse code, which is more energy efficient but requires more neurons to cover a space.  The dense code would approach $O(D\log N)$ total neurons, with systems with fewer rings approaching $O(DN)$ total neurons and with a correspondingly lower number of spikes.

\subsection{Density model}
One alternative to tracking the particles independently is to keep track of the density of particles at every location and randomly move walkers.  The main advantage of a particle density approach is that the complexity of the spatial graph is independent of the number of walkers.  While a density representation is the equivalent of the particle method in terms of producing estimated density distributions at different times, path dependent statistics are not readily available.  Instead, they must be decoded from the timing of the spikes.  This can impact some application, such as estimating the prices of certain path-dependent financial options.

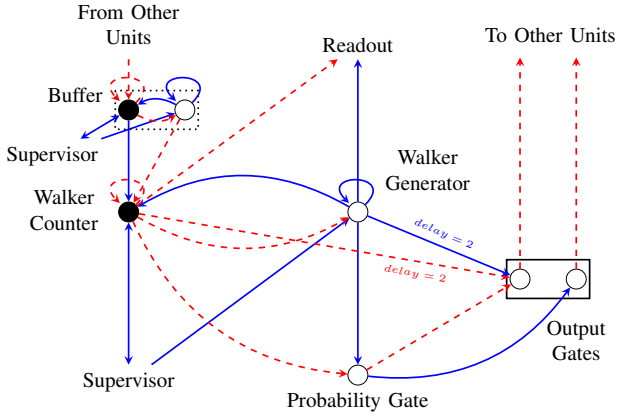
\begin{figure}
\scalebox{0.8}{
\begin{tikzpicture}[->,>=stealth,on grid,node distance=1.5in,inhibitory/.style={red, dashed},excitatory/.style={blue},persistent/.style={fill}]
\node[label=left:{\begin{tabular}{c}Walker\\ Counter\end{tabular}}] (counter) [draw,circle,persistent] at (0,0) {};
\node[label=above right:{\begin{tabular}{c} Walker\\ Generator\end{tabular}}] (generator) [draw,circle, right of=counter] {};
\node (readout) [above of=generator, above=-.5in] {Readout};
\node[label=below:{Probability Gate}] (probability) [draw, circle, below of=generator, below=-.5in] {};
\node (output) [above right of=probability,above=-.5in, circle, draw] {};
\node (output2) [right of=output, right=-1.2in, circle,draw] {};
\node (buffer) [above of=counter, above=-0.9in,draw, circle,persistent] {};
\node (bufferLabel) [above left=0.1in and 0.35in of buffer] {Buffer};
\node (bufferControl) [right of=buffer, right=-1.2in, draw, circle] { };
\node (bufferSupervisor) [below left=0.3in and 0.5in of buffer] {Supervisor};
\draw (bufferSupervisor) edge[<->, thick, excitatory] (buffer);
\draw (bufferSupervisor) edge[->, thick, excitatory] (bufferControl);
\draw[thick,dotted] ($(buffer.north west)+(-0.1,0.2)$)  rectangle  ($(bufferControl.south east)+(0.1,-0.2)$);
\draw[thick] ($(output.north west) + (-0.1, 0.2)$) rectangle ($(output2.south east) + (0.1, -0.2)$);
\draw (buffer) edge[->, thick, bend right, inhibitory]  (bufferControl);
\draw (bufferControl) edge[->, thick, bend right, excitatory] (buffer);
\draw (bufferControl) edge[->, thick, inhibitory] (counter);
\draw (bufferControl) edge[thick,loop, excitatory] (bufferControl);
\draw (buffer) edge[->, thick, inhibitory, loop] (buffer);
\node (input) [above of=counter, above=-.5in] {\begin{tabular}{c}From Other\\ Units\end{tabular}};
\node (supervisor) [below of=counter, below=-.5in] {Supervisor};
\draw (supervisor) edge[->, thick, excitatory] (generator);
\draw (supervisor) edge[<->, thick,excitatory] (counter);
\draw (generator) edge[->, thick, excitatory] (probability);
\draw (probability) edge[->, thick,excitatory, bend right] (output2);
\draw (probability) edge[->, thick, inhibitory] (output);
\draw (generator) edge[->, thick, bend right, excitatory] (counter);
\draw (counter) edge[->, thick, inhibitory] node[below,sloped,pos=0.75] {\tiny $delay= 2$} (output);
\draw (generator) edge[->, thick, excitatory] (readout);
\draw (counter) edge[->, thick, dashed, inhibitory] (readout);
\draw (counter) edge[->, thick, bend right, inhibitory]  (generator);
\draw (counter) edge[->, thick, bend right, inhibitory] (probability);
\draw (generator) edge[thick,loop, excitatory] (generator);
\draw (generator) edge[thick, excitatory] node[above,sloped] {\tiny $delay= 2$} (output);
\node (hidden) [above of=output] {};
\draw (output) edge[->, thick, inhibitory] (hidden);
\draw (counter) edge[thick, loop, inhibitory] (generator);
\draw (input) edge[thick, inhibitory] (buffer);
\draw (buffer) edge[thick, excitatory] (counter);
\node [below of=output2, below=-1.3in] {\begin{tabular}{c}Output\\ Gates\end{tabular}};
\node (hidden2) [above of=output2] {};
\draw (output2) edge[thick, inhibitory] (hidden2);
\node [above right=.1in and .2in of hidden] {To Other Units};

\end{tikzpicture}}
\caption{Schematic of a two-neighbor unit.    Circles represent neurons where empty represents complete decay ($Decay=1$) and filled represents no decay ($Decay=0$). Blue lines represent excitatory connections;  Red dashed lines are inhibitory. Weights are $\pm 1$ and delay is $delay=1$ unless marked.  Thresholds are all $0.5$ except for the filled (no decay) neurons that have threshold $0$.  The buffer circuit is optional and only used in synchronized mode,where all walkers take the same number of steps.  `Supervisor' generically refers to neurons used to moderate the walking process. \label{UnitSchematic}}
\end{figure}

As in the particle model, we need to either discretize a continuous space or equivalently assume that the markov process exists on a graph.  For each node on the graph, we instantiate a spiking circuit which we call a \textit{unit}.  A schematic of a two-neighbor unit is pictured in Fig.~\ref{UnitSchematic}.  Within a component there exists several key components:
\begin{enumerate}
\item \textit{Walker Counter~}  The walker counter is a simple neuron with threshold $0$ and contains running count of the number of walkers at a given node.  Walkers are passed from unit to unit by spikes with negative weight (inhibitory signal).  Hence, a sub-threshold potential of $-5$ corresponds with $5$ walkers being at the corresponding node.
\item \textit{Walker Generator~}  The walker generator is a self-excitatory neuron that `counts' out the walkers stored in the walker counter.  After being initiated by a separate supervisory signal, the walker counter sends positively weighted spikes to the walker counter, until the walker counter hits its threshold.  At this point, all walkers have started their next transition and inhibition from the walker counter halts the walker generator.
\item \textit{Probability Gate and Output Gate~} This group of neurons interacts with the output gates to ensure that each walker is sent to exactly one other unit, weighted by the specified probabilities.  More specifically, a tree of neurons subdivides (through selective excitation and inhibition) the potential outputs according to conditional probabilities. In Fig.~\ref{UnitSchematic}, the unit only has two neighbors and so only one neuron is needed for the random draw. 
\item \textit{Buffer~} The buffer is an optional component for synchronized operation. Without the buffer, the walkers may each take a different number of steps.  By incorporating a buffer, the walkers are first stored in the counter, sent to buffers of neighbor units, and then flushed from the buffers into the counters.  Structurally, the buffer contains a counter and generator neuron.
\end{enumerate}
The readout provides a mechanism for monitoring the simulation by observing the spikes being generated by the walker generator.  This could be useful for auxiliary computation or in hardware systems where sub-threshold potentials are unobservable. All the simulations that follow were performed on a software simulator and so, since we can directly access the sub-threshold potentials, a readout mechanism was not needed.

A simulation using this density model is performed as a series of manually or automatically triggered tasks.  Initially, current injection is used to place walkers at the desired initial position.  Then, walkers are counted and distributed by sending an excitatory signal to the walker counter and walker generator.  This automatically sends walkers to neighboring nodes via the the probability gate and output gate.  We connect a `walks complete' neuron downstream of the walker counters so that we know when all the walkers have been distributed.  If the units use synchronization buffers, the buffers are cleared in the same way via an excitatory signal.  Likewise, when the buffer is flushed, we use the resulting excitatory signal to trigger the next simulation timestep (i.e.~ the walkers take their next step).  Referring to the terminology of the particle model, the stochastic process occurs within the probability gates,   the spatial location is stored in the potentials of the walker counters, as each unit has a location, and the action circuit is a combination of the walker generator and the output gates.  

This density-based approach allows for the neuron requirement to be tied only to the size of the underlying space/graph and not to the number of walkers.  Overall, the neuron cost for a $n$-node graph is $O(n)$ assuming the number of neighbors for any node is much smaller than the total number of nodes.  The runtime is dependent on the number and distribution of walkers. The time taken to evaluate one simulation timestep is asymptotically linearly proportional to the largest number of walkers at a node.  

We note that for this construction, we assume that the underlying neuron model is capable of stochastic firing.  That is, after a threshold potential is exceeded, the neuron spikes according to the draw of a random number.  This stochastic model is representative of currently available neuromorphic hardware.  However, with more advanced neuron models, such as one that supports stochastic synapses (i.e.~spikes are sent to post-synaptic neurons according to independent random draws) could allow for simplified circuits.
\section{Simulations}
\subsection{Results of particle simulations}
\begin{figure}
\begin{center}
\includegraphics[width=.45\textwidth]{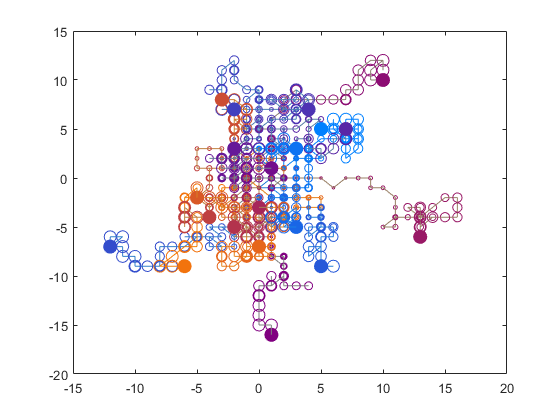}
\end{center}
\caption{Random walk of 20 particles over 100 time steps in free space originating at the origin, with a balanced probability of moving in either dimension equal to 0.25.  Each particle used rings of size 5, 7, and 11 neurons.  Steps are represented by progressively larger circles, with the solid dot representing the end location.}\label{Particle_20}
\end{figure}

\begin{figure}
\begin{center}
\includegraphics[width=.45\textwidth]{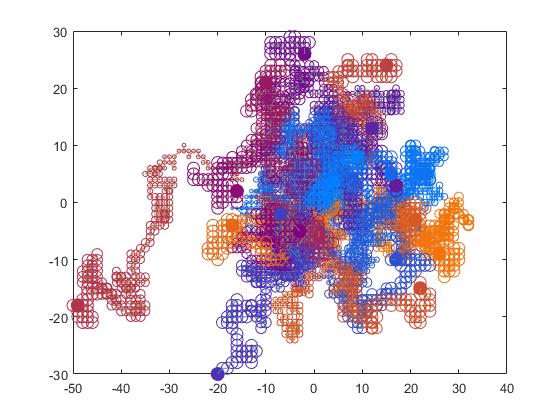}
\end{center}
\caption{Random walk of 20 particles over 1000 time steps in free space originating at the origin, with a balanced probability of moving in either dimension equal to 0.25.  Each particle used rings of size 5, 7, and 11 neurons.}\label{Particle_20_1000}
\end{figure}

First, we demonstrate the particle method by showing a random walk in free space.  Fig.~\ref{Particle_20} illustrates the appropriate random trajectories of the particles over a 100 time steps.  Fig.~\ref{Particle_20_1000} shows a longer time course, with particles moving for 1000 time steps.

\begin{figure}
\begin{center}
\includegraphics[width=.45\textwidth]{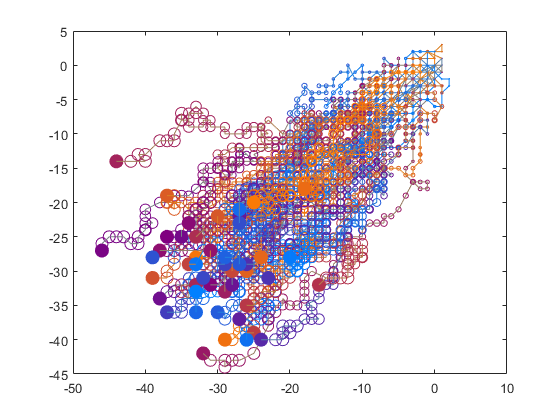}
\end{center}
\caption{Random walk of 50 particles over 200 time steps in free space originating at the origin, with a weighted probability (of 20\%) of moving in the negative direction in each dimension, versus 5\% of moving in the positive direction.  Each particle used rings of size 5, 7, and 11 neurons.}\label{Particle_weighted}
\end{figure}

Next, we illustrate how more complex walks can be examined, such as non-uniform probabilities.  Fig.~\ref{Particle_weighted} shows a case where random movement is biased heavily in the negative direction, with the movement of the particles drifting as a population towards the bottom left.  

\begin{figure}
\begin{center}
\includegraphics[width=.45\textwidth]{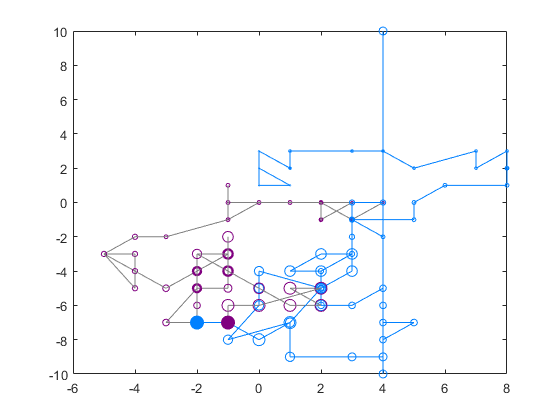}
\end{center}
\caption{Random walk of 2 particles over 50 time steps in free space originating at the origin, with a balanced probability of moving in either dimension equal to 0.25.  Each particle used rings of size 3 and 7.  Note the misencoding of the blue particle due to reaching the capacity of the code.}\label{Particle_2_err}
\end{figure}

One key limitation of the modular method described above is its behavior when a walker's position exceeds the precision of the neural circuit.  Because the modular code described above has a finite capacity, eventually particles in a free space will drift beyond the provided spatial resolution, wrapping around the space as if it is a torus.  An example of this is shown in Fig.~\ref{Particle_2_err}, wherein a small modular code (rings of size 3 and 7) led to a perceived jump of the particle from position $-10$ to $+10$.  

\begin{figure*}[h!]
\includegraphics[width=\textwidth]{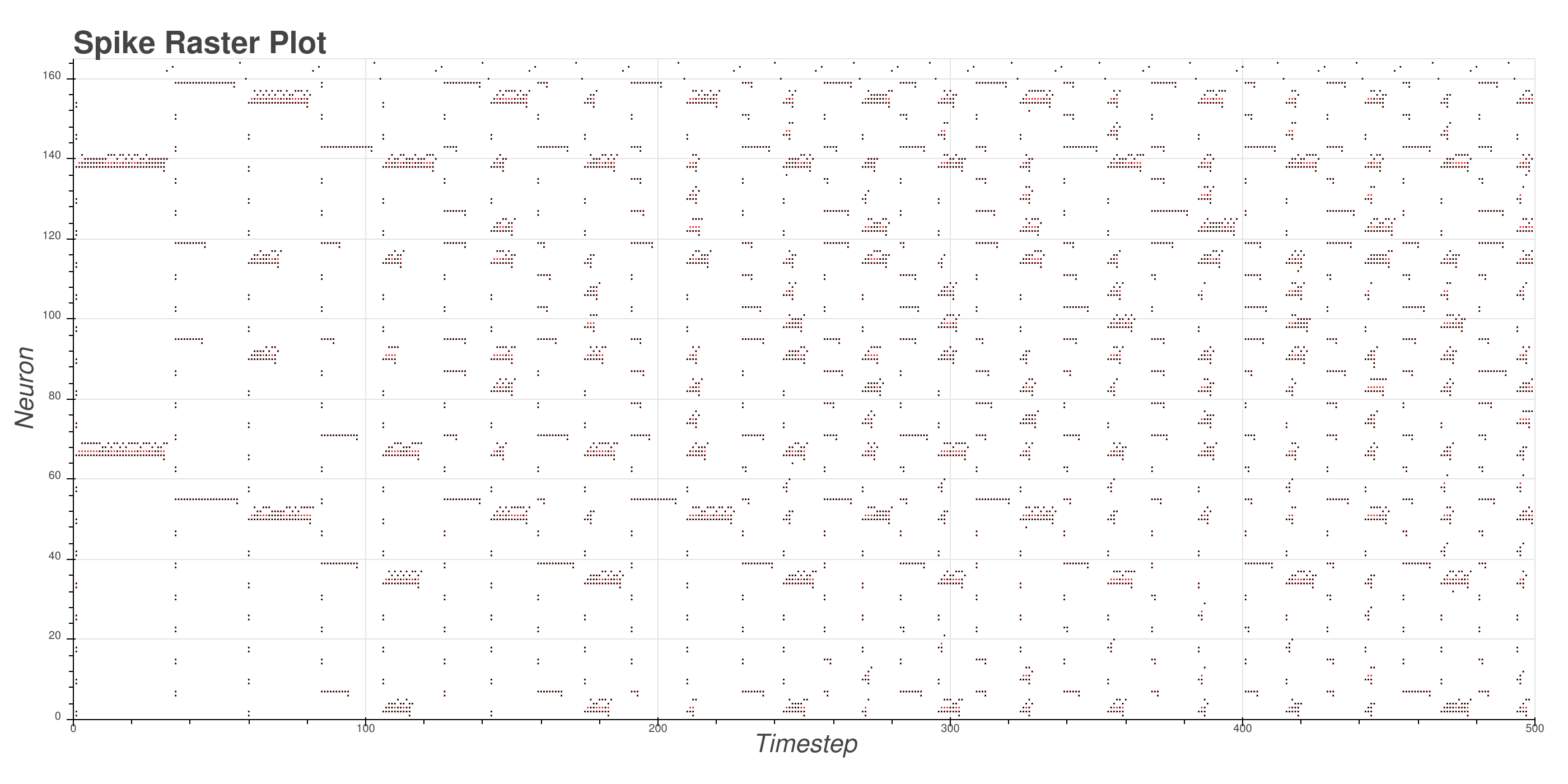}
\caption{The spike raster plot for a one-dimensional random walk, with $50\%$ probability in both up and down directions.  Walkers begin on units $10$ and $13$.  Black dots represent spike events; red dots represent failed probability calls (i.e.~neuron threshold is met, but the neuron does not spike due to stochasticity).  Neurons are grouped by unit, but are not sorted.  Each simulation time step requires fewer computational time steps as walkers become more diffuse.\label{SpikeRaster}}
\end{figure*}

\begin{figure*}[b!]
\includegraphics[width=\textwidth]{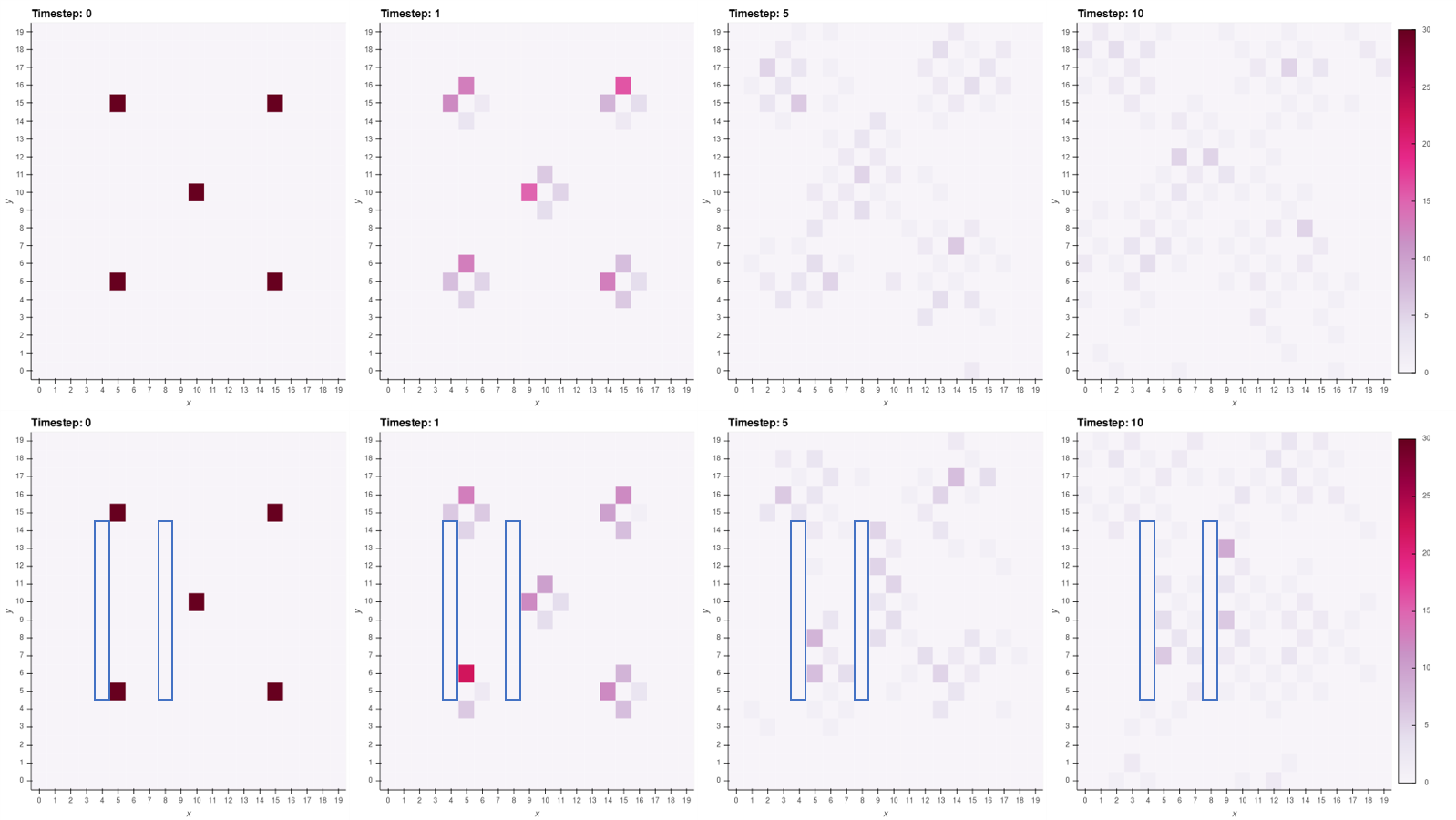}
\caption{Sample walker distributions from two separate two-dimensional experiments.  In the first row,  $30$ walkers start on units $(5,5)$, $(5,15)$, $(10, 10)$, $(15, 5)$, $(15,15)$; directions up, down, left, right have probabilities $35\%$, $35\%$, $15\%$, $15\%$ respectively. In the second row, the simulation has the same setup, except there are two obstacles (highlighted in blue).  The walkers have $0$ probability to enter the highlighted areas; the probability to enter a wall is distributed evenly to the perpendicular directions.  }
\label{2D-density}
\end{figure*}

\subsection{Results of density simulations}
\begin{figure}
\begin{center}
\includegraphics[width=.45\textwidth]{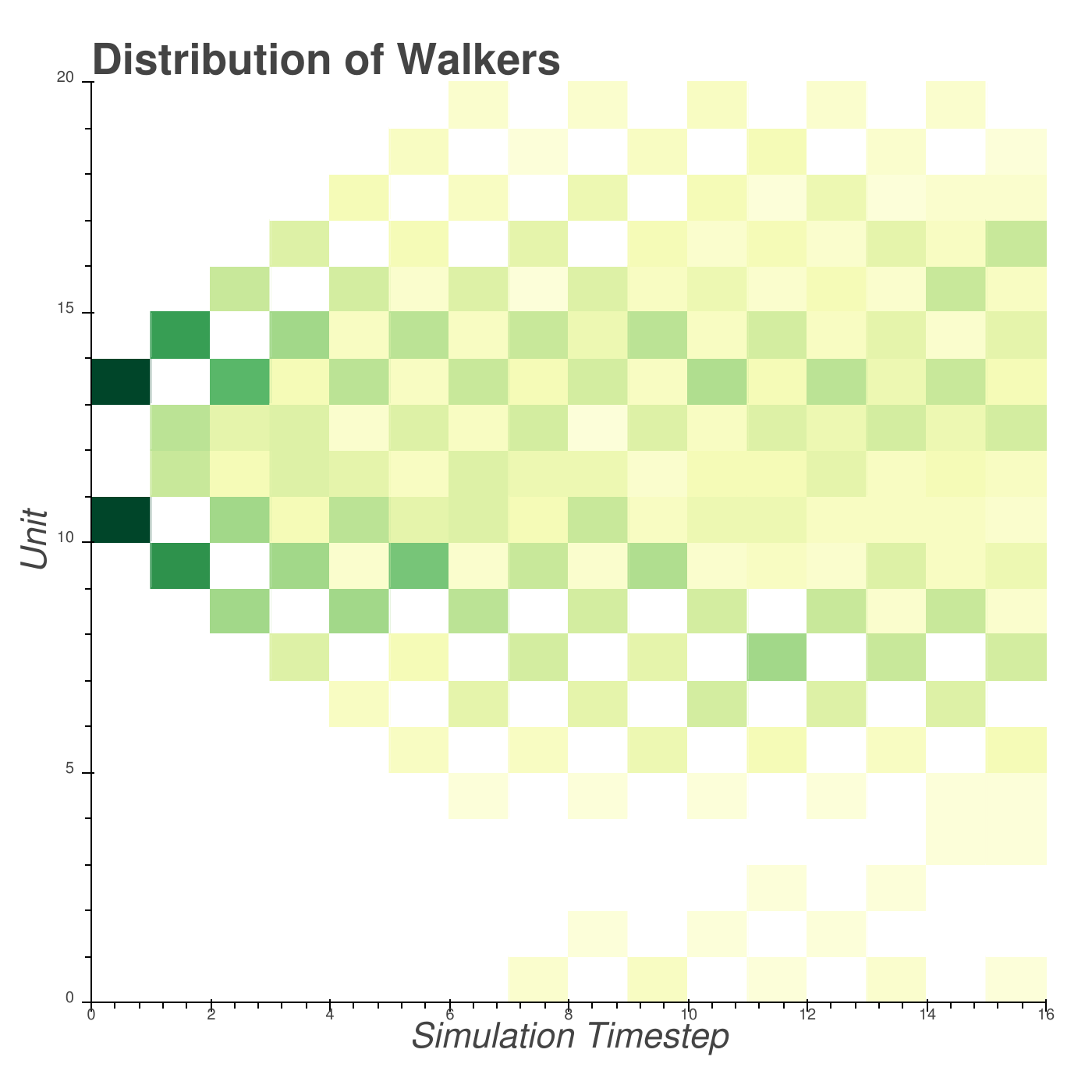}
\end{center}
\caption{ Walker distribution over time for a one-dimensional random walk, with $50\%$ probability in both up and down directions.  Walkers begin on units $10$ and $13$.  Units are arranged spatially in a cycle so that walkers can `wrap around.'\label{1dWalkerDistrubtion}}
\end{figure}

To examine the density model, we first explored a one-dimensional space where nodes are connected in a cycle, with transitions to adjacent nodes having a $50\%$ probability.  Pictured in Fig.~\ref{1dWalkerDistrubtion} is the distribution of walkers with units $10$ and $13$ being initialized with $30$ walkers each.  Only the Markov simulation time is shown.  The checkerboarding seen is a result of the fact that each node is connected only to two neighbor nodes, and any given walker must move to one of these two options.  

The corresponding spike raster for the 1D case is shown in Fig.~\ref{SpikeRaster}.  As expected, the walkers tend towards a uniform distribution.  Fig.~\ref{SpikeRaster} also illustrates how time is treated differently the density method.  The amount of simulation time required to model one time step of the system evolution is non-deterministic, requiring enough time to potentially move all particles within any given location.  Because the system progresses forward synchronously for spatial locations, the model time required to advance is dependent on the distribution of walkers.  Because we are modeling a random walk, most simulations will drift towards more diffuse distributions, requiring progressively less time to simulate the model.  

\begin{figure}
\begin{center}
\includegraphics[width=.45\textwidth]{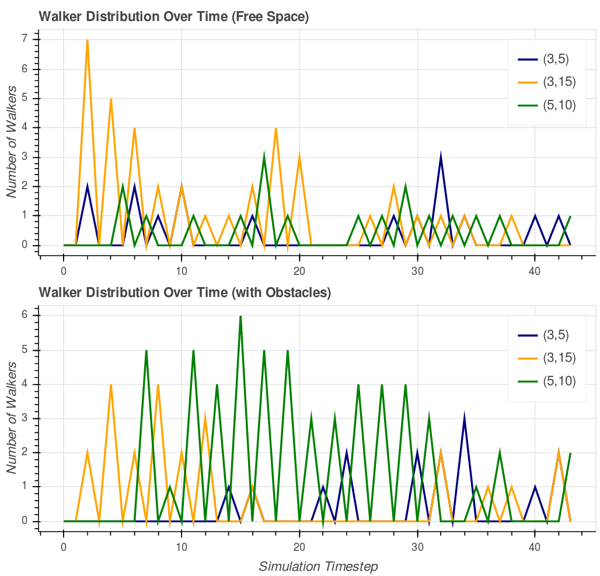}
\end{center}
\caption{ Plotted are the number of walkers over time at three different locations.  The top graph corresponds to the top row of Fig.~\ref{2D-density}, as the bottom graph does with the bottom row.  The evolution of the walker distribution is affected by the additional obstacles.\label{timeseries}}
\end{figure}

Fig.~\ref{2D-density} illustrates two time-courses of the density model in two dimensions on a torus.  The top half shows an instance where the probabilities are uniform across the space (though weighted towards the up and left directions).  The bottom half explores a case where walkers are prevented from entering two disjoint obstacles.  In Fig.~\ref{timeseries}, we plot the walker density at three different locations.

\section{Conclusions}
In this paper, we have demonstrated that small-scale neural circuits can efficiently and scalably implement random walk simulations.  While this paper does not examine other aspects of stochastic process models that are critical for many applications, such as complex boundary conditions and interactions between particles, the models in this paper are designed to be extended towards such considerations.  
 
Notably, the two models of random walks shown here are functionally equivalent, but each offer advantages under particular circumstances.  For instance, the number of neurons required for density method scales with spatial resolution, and the number of particles being modeled is dynamically accounted for in the time required for the model to run.  This configuration may thus be well-suited for neuromorphic systems whose neurons are capped at a fixed level whereas the time a simulation can be run is flexible.  Thus the number of particles can be tuned to achieve the statistical significance demanded by an application.    Alternatively, the particle method models each walker independently, thus the time for a simulation to run is independent of the number of walkers so long as there are sufficient neurons to represent the requisite spatial resolution within each neuron.

There are several reasons beyond scaling that one method may be preferable to the other.  While perhaps not as obvious, the paths taken by individual particles are preserved within the spike timings of the density method; however, the behavior of individual paths is directly retrievable from the particle methods.  This is of use in models of certain path-dependent financial options for instance \cite{goldman1979path}.  On the other hand, for many applications the density of walkers at a given spatial location and time is the critical output of stochastic process models.  The density method by its nature provides an estimation of the density at all locations of the space at all times, whereas the particle method would require a subsequent integration of information from all of the independent circuits.

Finally, the two models here each offer compelling potential advantages on different neuromorphic platforms, such as the IBM TrueNorth chip \cite{merolla2014million}, Intel's Loihi chip \cite{davies2018loihi}, Sandia's STPU archiecture \cite{hill2017spike}, and the Manchester SpiNNaker platform \cite{furber2013overview}.  The mapping of these algorithms to spiking neuromorphic systems will be a subject of a future study.  However, we anticipate that these algorithms should map well to these and other platforms, as the highly parallel nature of random walk processes makes them well suited for neuromorphic architectures.  We conclude by highlighting the point that the efficient implementation of a strictly numerical process such as the random walk on neuromorphic hardware would represent a major new capability for systems generally designed to implement tasks such as neural processing and machine learning. 

\section*{Acknowledgment}
This research was funded by the Laboratory Directed Research and Development program at Sandia National Laboratories.  Sandia National Laboratories is a multi-mission laboratory managed and operated by National Technology and Engineering Solutions of Sandia, LLC., a wholly owned subsidiary of Honeywell International, Inc., for the U.S.~Department of Energy's National Nuclear Security Administration under contract DE-NA0003525.

\bibliographystyle{IEEEtran}
\bibliography{bibliography}

\end{document}